\documentclass[journal]{IEEEtran}
\ifCLASSINFOpdf
\usepackage[pdftex]{graphicx}
\else
\usepackage[dvips]{graphicx}
\fi
\usepackage{amsmath}
\usepackage{mathptmx}
\usepackage{amssymb}
\usepackage{amsthm}
\usepackage{comment}
\usepackage[ruled,vlined]{algorithm2e}
\usepackage{hyperref}
\usepackage{caption}
\usepackage{subcaption}
\usepackage{etoolbox}
\usepackage{threeparttable}
\AtBeginEnvironment{algorithm}{\SetArgSty{textrm}} % <-'Flag' will not be written in italic

\begin{document}

	%
	% paper title
	% Titles are generally capitalized except for words such as a, an, and, as,
	% at, but, by, for, in, nor, of, on, or, the, to and up, which are usually
	% not capitalized unless they are the first or last word of the title.
	% Linebreaks \\ can be used within to get better formatting as desired.
	% Do not put math or special symbols in the title.
	\title{Shuffle and Learn: Minimizing Mutual Information for Unsupervised Hashing}
	%
	%
	% author names and IEEE memberships
	% note positions of commas and nonbreaking spaces ( ~ ) LaTeX will not break
	% a structure at a ~ so this keeps an author's name from being broken across
	% two lines.
	% use \thanks{} to gain access to the first footnote area
	% a separate \thanks must be used for each paragraph as LaTeX2e's \thanks
	% was not built to handle multiple paragraphs
	%
	
	\author{\IEEEauthorblockN{Fangrui~Liu\IEEEauthorrefmark{1}~and~Zheng, Liu\IEEEauthorrefmark{2}}\\
		\IEEEauthorblockA{Faculty of Applied Science,
		University of British Columbia\\
		Email: \IEEEauthorrefmark{1}fangrui.liu@ubc.ca,
		\IEEEauthorrefmark{2}zheng.liu@ubc.ca}}

	% The paper headers
	%\markboth{Transactions on Neural Networks and Learning Systems,~Vol.~X, No.~X, November~2020}%
	%{Shell \MakeLowercase{\textit{Fangrui et al.}}: Minimizing Mutual Information for Unsupervised Binary Representation}
	% The only time the second header will appear is for the odd numbered pages
	% after the title page when using the twoside option.
	% 
	% *** Note that you probably will NOT want to include the author's ***
	% *** name in the headers of peer review papers.     ***
	% You can use \ifCLASSOPTIONpeerreview for conditional compilation here if
	% you desire.

	% If you want to put a publisher's ID mark on the page you can do it like
	% this:
	%\IEEEpubid{0000--0000/00\$00.00~\copyright~2015 IEEE}
	% Remember, if you use this you must call \IEEEpubidadjcol in the second
	% column for its text to clear the IEEEpubid mark.

	% use for special paper notices
	%\IEEEspecialpapernotice{(Invited Paper)}

	% make the title area
	\maketitle
	
	% As a general rule, do not put math, special symbols or citations
	% in the abstract or keywords.
		\begin{abstract}
		Unsupervised binary representation allows fast data retrieval without any annotations, enabling practical application like fast person re-identification and multimedia retrieval. It is argued that conflicts in binary space are one of the major barriers to high-performance unsupervised hashing as current methods failed to capture the precise code conflicts in the full domain. A novel relaxation method called \textit{Shuffle and Learn} is proposed to tackle code conflicts in the unsupervised hash. Approximated derivatives for joint probability and the gradients for the binary layer are introduced to bridge the update from the hash to the input. Proof on $\epsilon$-Convergence of joint probability with approximated derivatives is provided to guarantee the preciseness on update applied on the mutual information. The proposed algorithm is carried out with iterative global updates to minimize mutual information, diverging the code before regular unsupervised optimization. Experiments suggest that the proposed method can relax the code optimization from local optimum and help to generate binary representations that are more discriminative and informative without any annotations. Performance benchmarks on image retrieval with the unsupervised binary code are conducted on three open datasets, and the model achieves state-of-the-art accuracy on image retrieval task for all those datasets. Datasets and reproducible code are provided\footnote{ \url{https://github.com/mpskex/Minimizing-Mutual-Information}}.
	\end{abstract}
	
	% Note that keywords are not normally used for peerreview papers.

	% For peer review papers, you can put extra information on the cover
	% page as needed:
	% \ifCLASSOPTIONpeerreview
	% \begin{center} \bfseries EDICS Category: 3-BBND \end{center}
	% \fi
	%
	% For peerreview papers, this IEEEtran command inserts a page break and
	% creates the second title. It will be ignored for other modes.
	\IEEEpeerreviewmaketitle

	\section{Introduction}
	% The very first letter is a 2 line initial drop letter followed
	% by the rest of the first word in caps.
	% 
	% form to use if the first word consists of a single letter:
	% \IEEEPARstart{A}{demo} file is ....
	% 
	% form to use if you need the single drop letter followed by
	% normal text (unknown if ever used by the IEEE):
	% \IEEEPARstart{A}{}demo file is ....
	% 
	% Some journals put the first two words in caps:
	% \IEEEPARstart{T}{his demo} file is ....
	% 
	% Here we have the typical use of a "T" for an initial drop letter
	% and "HIS" in caps to complete the first word.
	
	% Application on Binary Representation
	\IEEEPARstart{B}{inary} representations are designed to be compact and informative. It is also called as data hashing, which is efficient on both computation and storage. High dimensional data like images and audio clips consumes large room on the disk, and it was almost impossible to index, search and understand on computers. With the advances in machine learning and pattern recognition, binary representation learned by deep neural networks takes smaller space on storage. Those learned binary codes can measure similarity and classify with considerable accuracy compared to the continuous counterpart. Generally, concrete supervision, \textit{eg.} class categories, is needed to achieve high performance on tasks like image retrieval with binary representation. However, in most realistic scenarios, annotations like labels are hard to obtain for a generic image retrieval application. Researches on unsupervised hashing algorithms leverage the representation learning with smaller code size and more flexibility on application.
	
	\begin{figure}
		\centering
		\begin{subfigure}[b]{0.475\linewidth}
			\centering
			\includegraphics[width=\linewidth]{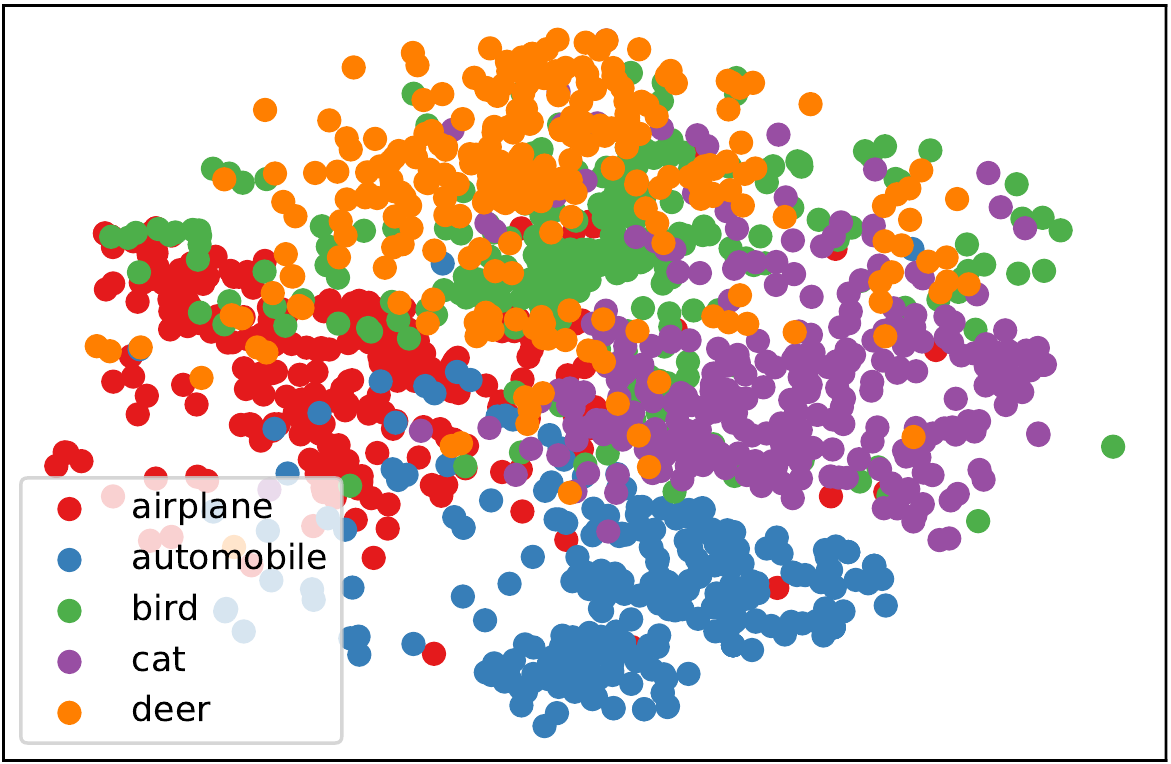}
			\caption{Without Relaxation}
			\label{fig: story a}
		\end{subfigure}
		\begin{subfigure}[b]{0.475\linewidth}
			\centering
			\includegraphics[width=\linewidth]{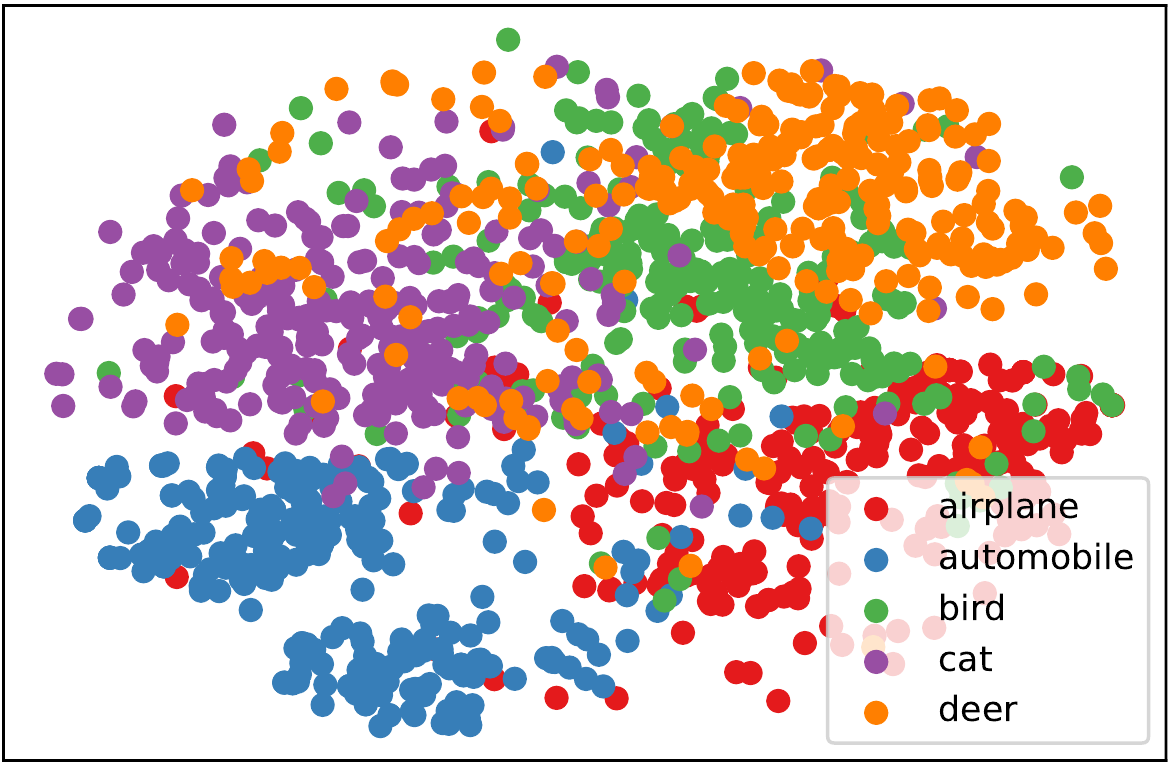}
			\caption{With \textit{Shuffle and Learn}}
			\label{fig: story b}
		\end{subfigure}
		\caption{\sc Visualized Effect on Minimizing Mutual Information over Binary Space}
		\label{fig:story}
	\end{figure}
	
	Unsupervised low dimensional binary representation learning, also known as unsupervised hashing, is the main topic in this paper. Binary hash boosts the speed on image searching, as the representation is more sparse than the continuous code. Even simple metrics like Hamming distance is effective to retrieve similar images in the domain with binary representations. And also other applications like cross-modal retrieval \cite{yu2016binary,jin2018deep}, Face identification \cite{tang2018discriminative} have proven that binary code is informative enough to distinguish samples in a large domain. On the other hand, unsupervised binary representations are ideal for unknown data. No prior knowledge like bounding boxes and labels is needed to train an unsupervised binary hashing algorithm. It is also capable of carrying sufficient information in application like multimedia retrieval \cite{xie2016unsupervised,wu2018unsupervised} without concern on annotations. In conclusion, across different domains, an unsupervised hash can be widely utilized in applications that require compact representation for unlabeled data.
	
	Unsupervised binary representation learning, unlike the supervised hash, is a challenging topic due to its strong sparsity and insufficiency on constraints with ground truth. Low dimensional binary space has fewer keys in code space to map the samples in the domain. It will drastically reduce the expressiveness by shrinking the value set from the real numbers to binary. To recapture the information, current methods borrow the concept of `similar' and `dissimilar' from the supervised hashing algorithms. Metrics like cosine similarity is applied to preserve the similarity of the binarized representation from the feature holds for the original data. Other constraints are also used to minimize the error between binary code and the learned continuous representation. Intuitively, learning unsupervised representations can be considered as learning a dictionary in the domain. Better representations usually use up more keys in the binary space. Hence, in this work, a novel constraint will be introduced which encourages binary code to diverge and fill up the binary space for the domain.
	
	Conventional unsupervised hashing suffers from inefficient use of the code space, which harms the performance on distinguishing samples in the domain. Samples will tend to share the same binary code which is incapable of identifying them in learned unsupervised hash database. Models always suffer from code conflict issue in Fig. \ref{fig: story a}, while models with proper relaxation will generalize better, separating samples in a wider range as shown in Fig. \ref{fig: story b}. Current methods \cite{heo2012spherical,gong2012iterative,liu2014discrete,su2018greedy,yang2019distillhash} only apply losses that maximize the similarity and consistency of feature, hidden code and the hashed code, which may push the solution into a local optimum. It will cause conflicts where samples that should be discriminated sharing the same or similar binary code during optimization. Therefore, proper relaxation needs to be proposed to help the network jump out of the local optimum and scatter representation better in the binary space.
	
	Mutual information is a good criterion to evaluate the correlation between random variables and can be used as a reference to optimize coding efficiency during training. Nevertheless, it is hard to directly obtain gradients with mutual information as both the binarization and joint probability block the gradients that update network parameters. Fortunately, a greedy binarization has already been proposed with code consistency loss \cite{su2018greedy}. Then the only challenge left is to bridge gradient on joint probability. To tackle that, an approximated derivative is proposed for joint probability to connect those gradients from the output to inputs, enabling back-propagation with mutual information over the learned binary representation.
	
	Our contribution can be summarized as:
	\begin{enumerate}
		\item An relaxation on unsupervised hash called \textit{Shuffle and Learn} is proposed to encourage less code conflict in the unsupervised hash by minimizing mutual information.
		\item Proof on the $\epsilon$-Convergence is provided to justify the effectiveness of minimizing mutual information with approximated derivatives on joint probability.
		\item Experiment result suggests that the proposed method achieves SOTA on image retrieval task with unsupervised binary representation (50.7\% on CIFAR-10(II), 71.5\% on NUS-Wide, 70.3\% on MS-COCO with 16-bits binary code)
	\end{enumerate}
	\par
	The paper will be delivered into five sections including the introduction. Related researches will be discussed in Section \ref{sec:related}, covering work on unsupervised binary representation, mutual information in binary representation learning and regularization on unsupervised deep clustering. The proposed method is introduced in Section \ref{sec:method}, coming along with how the gradients are bridged during mutual information minimization, $\epsilon$-Convergence proof on joint probability with approximated derivatives and the algorithm proposed as an application on mutual information minimization. Experiment results will be reported in Section \ref{sec:experiments}. At the end of this paper, we will conclude the contribution we made in this work in Section \ref{sec:conclusion}.
	
	\section{Related Work}
	\label{sec:related}
	
	There is much research on deep binary representation learning in the past decade. Supervised binary representation learning has been in trend for many years \cite{jin2018deep,tang2018discriminative,zhe2019deep,wu2019deep,do2019compact,shen2019embarrassingly,li2020weighted}. The proposed approach is related to three major research directions, which are unsupervised binary representation learning, mutual information in binary representation learning and regularization on deep clustering.
	
	\subsection{Unsupervised Binary Representation Learning}
	Unsupervised hashing methods always take features from sophisticated extractors like VGG \cite{simonyan2014very} or ResNet \cite{He_2016_CVPR} and learn a binary hash code which describes its semantic representation without any labels or hints. Current researches in unsupervised hashing can be categorized into two series. Discriminative networks constrained by unsupervised losses is one major direction in unsupervised hashing methods. Rotation invariant is considered to retain maximum semantic information from the binary representation \cite{lin2016learning}. Quantization loss and even distribution on the learned hash are also considered to keep the capability of the binary hash to the original continuous feature representation. A triplet loss \cite{huang2017unsupervised} is used to generate discriminative hash while keeping consistency and capability on the binarized hash. Samples are considered into positives and negatives when being compared to the original. Positive samples, which are generated with random rotation from the original, should be closer to the negatives that are randomly picked from the dataset. Also, the angular distance among samples provides evidence to similarity \cite{hu2017hashing} as the binary code only take the sign of features into its account. Furthermore, the relation can be learned with mild assumption \cite{yang2019distillhash}. A Similarity is predicted with Bayesian optimal classifier during the training process to find the distil data pair which is further used in hash learning. Alternatively, a graph can also describe feature similarity \cite{shen2018unsupervised}, guiding the network to search better hash for the data distribution. Samples are defined as vertices and the similarity are described as edges in the graph. For each iteration, the edges will be reinitialized with the current similarity on learned features.
	\par
	Generative methods are also popular in unsupervised binary representation learning. Adversarial learning seems to be effective on extracting binary hash code. BinGAN \cite{zieba2018bingan} introduces distance matching and entropy regularization with discriminator to learn how to hash effectively. Similarly, HashGAN \cite{ghasedi2018unsupervised} also adopted a generator-discriminator architecture to learn binary representation. Both even distribution and minimum entropy are considered in an adversarial learning framework. Variational Autoencoders are also effective in finding a proper hash function on datasets. DVB \cite{shen2019unsupervised} considers reconstruction on the learned bits, which can reflect the information that a hash can retain and also can be a criterion to maximize the representativity of the binarized representations.
	\par
	In conclusion, unsupervised binary representation learning always considers semantic representativity among the binary representations as well as fewer code conflicts in the full domain. Current unsupervised hashing methods are usually targeted at keeping semantic information in learned binary code, which may lead to local optimum during learning. However, they failed to relax the code conflicts when extracting unsupervised semantic hash. Those code collision in hashing models will also harm binary code's representativity in the binary space. Therefore, proper regularization is necessary to relax code conflicts when learning unsupervised hash.
	
	\subsection{Regularization on Unsupervised Deep Clustering}
	Unsupervised hashing is similar to deep clustering as each key in the code space can be interpreted as a cluster gathering similar samples for the learned feature. Regularization on deep clustering can be applied to unsupervised hash to improve code quality. Researches on deep clustering regularization convinced us that inter-class conflicts are crucial for good binary representation learning \cite{Zhao_2019_CVPR}. Relative entropy minimization is introduced to gather similar data into a rough cluster by reconstructing the original example from distorted data \cite{ghasedi2017deep}. Entropy regularization on embedding with adaptive weights is proposed to enhance the feature robustness in fuzzy k-mean clustering \cite{zhang2019deep}. Other regularizations on clustering, like a differentiable constraint on cluster size \cite{genevay2019differentiable} and structural regularization \cite{tang2020unsupervised}, are also introduced to avoid embedding conflicts in the continuous space. 
	\par
	Regularization is also important to eliminate code conflicts in the code space on deep clustering if we consider clusters as code in the binary space. However, deep clustering often works on continuous representation to form small groups without any supervision. Most inter-class collision regularization from deep clustering can not be directly applied to unsupervised binary representations.

	\begin{comment}
	\subsection{Mutual Information in Binary Representation Learning}
	
	Mutual information plays an important role in binary representation learning as it provides a correlation between groups in the domain. A mutual information loss is proposed for locality-sensitive hashing (LSH) and is proved to $f$-divergence of the framework with two-side approximation \cite{chen2019locality}, discriminating distributions better the simple Hamming distance in LSH scheme. Also, mutual information is introduced to divergence class-wise representation distribution in supervised binary representation learning \cite{cakir2017mihash,cakir2019hashing}. An approximated derivative is given to complete the chain rule.
	\par
	Generally, current researches on hashing with mutual information mainly focus on diverging local representation distribution by minimizing a class-wise or a locality-sensitive mutual information loss to the learned binary representation. Besides, such a criterion can also be applied on individual bits to regularize the correlation among each single binarized neurons if they are considered as identical independent events.
	\end{comment}
	
	\section{Shuffle and Learn}
	\begin{table}[ht]
		\centering
		\caption{\sc Table of Notation}
		\begin{tabular}{ l l }
			\hline
			Notation&Description\\
			\hline
			$B_i$&Output binary random variable on $i$-th position\\
			$P(B_i)$&Probability of when $B_i$ is true\\
			$H(B_i)$&Entropy of $B_i$\\
			$I(B_i;B_j)$&Mutual information between $B_i$ and $B_j$\\
			$\theta$&Network parameter\\
			$\eta$&Learning rate for parameter update\\
			$\epsilon$&Error on approximated joint probability\\
			$\Delta_i$&Update gradient on $P(B_i)$\\
			\hline
		\end{tabular}
		\label{tab: notation}
	\end{table}
	\label{sec:method}
	% explainning the mutual information is intuitive to optimize
	Mutual information plays an important role in binary representation learning as it provides a correlation between groups in the domain. Mutual information is introduced to divergence class-wise representation distribution in supervised binary representation learning \cite{cakir2017mihash,cakir2019hashing}. Conversely, the proposed \textit{Shuffle and Learn} tries to minimize mutual information between bit pairs in hash code, encouraging less code conflict in the learned unsupervised binary representation.
	\par
	Mutual information indicates the mutual dependencies between two random variables. The learned binary representation can also be considered as a group of random variables $P(B_i)$ for the $i$-th in $N$ bits. Intuitively, the correlation between arbitrary pair $\{B_i, B_j\}$ can be eliminated by minimizing the mutual information $I(B_i; B_j)$. Minimizing the mutual information can diverge the samples to use up the whole binary coding space. Generally, the unsupervised binary representation learning may suffer from local optimum, while mutual information minimization would help to jump out of it. Gradients from mutual information minimization guide the model to a more spare solution that provides better accuracy and discriminative representations to describe the original data.
	\par
	In this section, we will first discuss how those gradients are bridged the approximated gradient for joint probability. As the gradients are obtained from the approximated joint probability, we need to prove the approximation will eventually converge at the real magnitude of joint probability. At the end of this section, we will introduce an application of the proposed method, regularizing the unsupervised binary representation learning by minimizing the mutual information among bits it learned. Used notations are listed in Table \ref{tab: notation}.
	
	\subsection{Approximated Gradient for Joint Probability}
	% Evaluation on Mutual Information
	
	Minimizing mutual information is an optimization problem and gradients are required to update parameters. The partial derivative of $I(B_i;B_j)$ on $P(B_i, B_j)$ is easy to obtain but it is not trivial for the partial derivative of $P(B_i, B_j)$ on $B_i$. Therefore, we will mainly focus on bridging gradient for $P(B_i, B_j)$.
	\par
	Let $f$ be our differentiable learning function and $b_i$ be the $i$-th output among $N$ bits from function $f(~\cdot~;\theta)$ with parameter $\theta$. The target function is mutual information $I(B_i, B_j)$ which is going to be minimized. The partial derivative $\frac{\partial I(B_i, B_j)}{\partial \theta}$ is needed to minimize the observed mutual information. Then the overall partial derivative chain can be derived as Eq.~\eqref{eq:grad derive}.
	\begin{equation}
	\label{eq:grad derive}
	\frac{\partial I(B_i, B_j)}{\partial \theta} = \frac{\partial I(B_i, B_j)}{\partial P(B_i, B_j)}\frac{\partial P(B_i, B_j)}{\partial B_i}\frac{\partial B_i}{\partial f}\frac{\partial f}{\partial \theta}
	\end{equation}
	\par
	There are two parts that is `broken' in the derivative chain. One is the binary layer part $\frac{\partial b_i}{\partial f}$ and another is the joint probability part $\frac{\partial P(B_i, B_j)}{\partial b_i}$. For the binary part, a simple straight through gradient strategy \cite{su2018greedy} is applied to update the binary representation with the continuous gradient with a code constraint. Then the only problem would be the derivative on joint probability.
	\par
	Statistically, accumulating partial derivatives to a random variable from joint probability does not make sense. However, the gradient is required to link the output to inputs in differentiable learning functions. By introducing positive and negative association on binary random variable pairs, the gradient can be easily obtained to update according to the computed statistics.
	\par
	Positive association \cite{esary1967association} is proposed to describe a pair of variables $B_i$ and $B_j$ that satisfies $Cov[B_i, B_j] \geq 0$ where $Cov[B_i, B_j] = E[B_i, B_j] - E[B_i]E[B_j]$. Specifically, the expectation of a binary random variable is the probability when itself is positive, making it easier to find its lower bound for the joint probability. Hence, for a pair of positively associated binary variables, the upper bound can be easily derived as Eq.~\eqref{eq:pos asso major}.
	\begin{equation}
	\label{eq:pos asso major}
	P(B_i, B_j) \geq P(B_i)P(B_j)
	\end{equation}
	\par
	Negative association \cite{joag1983negative} is the opposite to its positive counterpart, where $Cov[B_i, \bar{B_j}] \leq 0$. We inherit the notation from previous derivation for clarification by assuming pair $\{B_i, B_j\}$ is in positive association. So that for a pair of negatively associated binary variables $\{B_i, \bar{B_j}\}$, the lower bound can be easily obtained as Eq.~\eqref{eq:neg asso major}.
	\begin{equation}
	\label{eq:neg asso major}
	P(B_i, \bar{B_j}) \leq P(B_i)P(\bar{B_j})
	\end{equation}
	\par
	With Eq.~\eqref{eq:pos asso major} and \eqref{eq:neg asso major} it is trivial to obtain the inequality for joint probability for a pair of positively associated binary random variables, as derived in Eq.~\eqref{eq:pos asso bound}.
	\begin{equation}
	\label{eq:pos asso bound}
	\begin{aligned}
	P(B_i, B_j) &\geq P(B_i)P(B_j)\\
	P(B_i, \bar{B_j}) &\leq P(B_i)-P(B_i)P(B_j)\\
	P(\bar{B_i}, B_j) &\leq -P(B_I)P(B_j) + P(B_j)\\
	P(\bar{B_i}, \bar{B_j}) &\geq 1-P(B_i)-P(B_j)+P(B_i)P(B_j)
	\end{aligned}
	\end{equation}
	\par
	Similarly, the negative will also provide reversed boundary for $P(B_i, B_j)$. A pair of associated binary code is either positively or negatively associated in statistics.
	Intuitively, each pair of independent binary code will be not associated when it is at the optimal solution subject to the mutual information. Our goal is to eliminate association among pairs of hash. When it is in its optimal solution, all inequalities in Eq.~\eqref{eq:pos asso bound} will turn into equalities, or it will have a positive or negative error from the joint to multiplication of marginals. To notate the error, we define a slack variable $\epsilon$ to describe the upper or lower limit of the error. Then we can use this slackness to prove the $\epsilon$-Convergence on this approximation.
	\par
	Joint probability for associated pair then can be rewritten as Eq.~\eqref{eq:bound slack} with the slack variable $\epsilon$:
	\begin{equation}
	\label{eq:bound slack}
	P(B_i, B_j)=P(B_i)P(B_j)-\epsilon\geq P(B_i)P(B_j)
	\end{equation}
	where the slack variable $\epsilon$ is bounded, which means it will never go infinity as both the joint $P(B_i,B_j)$ and $P(B_i)P(B_j)$ are bounded. Both the upper bound and lower bound of joint probability can be obtained by mixing positive and negative association condition. The slack variable $\epsilon$ will converge and squeeze the joint probability to be independently multiplied. It will stagger around zero as the variable pairs will switch between positive and negative association frequently. The derivative obtained by the expanded value can provide an approximated gradient that establishes a relationship between joint probability and the output binary representation. In the next section, we will discuss the convergence on the optimization with approximated derivatives with the slack variable $\epsilon$.
	\par
	With the slackness $\epsilon$, the derivative can be trivially derived as Eq.~\eqref{eq:derivation derivatives}.
	\begin{equation}
	\label{eq:derivation derivatives}
	\frac{\partial P(B_i,B_j)}{\partial B_i}\approx\frac{\partial P(B_i)P(B_j)}{\partial B_i}=P(B_j)
	\end{equation}
	where $\frac{\partial P(B_i)}{\partial B_i}=\frac{1}{N}$ for statistics over $N$ samples as the unconditional probability density function of binary variable $B_i$ can be considered as summation over positive samples. The derivatives of the rest conditions are stated in Eq.~\eqref{eq:approx derivative}. And the gradient can be calculated according to those approximated derivatives.
	\begin{equation}
	\label{eq:approx derivative}
	\begin{aligned}
	\frac{\partial P(B_i,\bar{B_j})}{\partial B_i}&\approx\frac{\partial P(B_i)P(\bar{B_j})}{\partial B_i}=1-P(B_j)\\
	\frac{\partial P(B_i,\bar{B_j})}{\partial B_i}&\approx\frac{\partial P(\bar{B_i})P(B_j)}{\partial \bar{B_i}}=-P(B_j)\\
	\frac{\partial P(\bar{B_i},\bar{B_j})}{\partial B_i}&\approx\frac{\partial P(\bar{B_i})P(\bar{B_j})}{\partial \bar{B_i}}=-\big(1-P(B_j)\big)
	\end{aligned}
	\end{equation}
	\par
	From a analytic perspective, compensating the bi-variant function with a function of one variable is an over-acting on the gradient. The slack variable will bounce between the optimum which would cause diverge with large step sizes. However, it will eventually provide more precise gradient with the convergence of $\epsilon$. Further discussion will be covered in next section and we proved that with small step sizes the slack variable will converge to 0 at last.
	\par
	The final gradient will accumulate the gradients on $B_i$ with respect to other bits $\{B_{j_1},...B_{j_N}\}$, and the actual accumulated joint probability will be used to compute loss when the network feeds forward.
	\subsection{ $\epsilon$-Convergence of Joint Probability Approximation}
	\label{subsec:e-convergence}
	The convergence of joint probability approximation needs to be proved to ensure accurate update before being applied during optimization. In the previous section, we introduced a slack variable $\epsilon$ which defines the upper and lower bound of the difference between the estimated and approximated joint probability. 
	\par
	In Eq.~\eqref{eq:derivation derivatives}, the partial derivative is obtained using the approximated joint probability $P(B_i)P(B_j)-\epsilon$. We will introduce a learning step-size $\eta^t$ to formulate the updated joint probability with the approximated gradient. Then a single iteration at $i$-th step can be derived as Eq.~\eqref{eq:update single}.
	\begin{equation}
	\label{eq:update single}
	\begin{aligned}
	P^{t+1}(B_i, B_j) &= P^t(B_i, B_j) - \eta^t\frac{\partial P^t(B_i, B_j)}{\partial P^t(B_i)}\\
	&=P^t(B_i, B_j) -\eta^t P^t(B_j)
	\end{aligned}
	\end{equation}
	By introducing the slack variable $\epsilon^t$ and the gradient $\Delta^t_i$ and $\Delta^t_i$ on $P^t(B_i)$ and $P^t(B_j)$ respectively for $t$-th iteration, we can replace $P^t(B_i, B_j)$ with $P^t(B_i)P^t(B_j)+\epsilon$ according to Eq.~\eqref{eq:bound slack} and also substitute $P^t(B_i) - \Delta^t_{P_(B_i)}$ for $P^{t+1}(B_i)$. Then the update can be trivially derived as Eq.~\eqref{eq: update slack derive 1}.
	\begin{equation}
	\label{eq: update slack derive 1}
	\begin{aligned}
	&-\Delta^t_jP(B_i)-\Delta^t_iP(B_j) +\Delta^t_i\Delta^t_j+\epsilon^{t+1}\\
	&\hskip1.2cm =\epsilon^t -\eta^t P^t(B_j)
	\end{aligned}
	\end{equation}
	Obviously, we can easily obtain the error difference on slack variable $\epsilon$ from iterations to iterations by deriving the update function to the joint probability with approximation. The error difference $\epsilon^{t+1} - \epsilon^t$can be illustrated as Eq.~\eqref{eq:error difference}.
	\begin{equation}
	\label{eq:error difference}
	\Delta^t_iP^t(B_j)+\Delta^t_jP^t(B_i)-\Delta^t_i\Delta^t_j-\eta^t P^t(B_j)
	\end{equation}
	\par
	To prove the convergence, we add all differences to form a series in $t$. Then we can build a relationship between $\epsilon^T-\epsilon^1$ and the series described in Eq.~\eqref{eq:error difference}. As both the learning rate converges to 0 and the probability is non-negative and bounded, we can imply that series $\sum_t^\infty{-\eta^t P(B_j)}$ is convergent. Then all we need to do is to prove series in Eq.~\eqref{eq:series derivation} is convergent.
	\begin{equation}
	\label{eq:series derivation}
	\begin{aligned}
	&\epsilon^T-\epsilon^1\\
	&=\Delta^t_i\big(P^t(B_j)-\frac{1}{2}\Delta^t_j\big)+\Delta^t_j\big(P^t(B_i)-\frac{1}{2}\Delta^t_i\big)\\
	&=\Delta^t_i\frac{P^t(B_j)+P^{t-1}(B_j)}{2}+\Delta^t_j\frac{P^t(B_i)+P^{t-1}(B_i)}{2}
	\end{aligned}
	\end{equation}
	Since $\frac{P^t(B_j)+P^{t-1}(B_j)}{2}\in[0,2]$, we can split and expand the component in Eq.~\eqref{eq:series derivation} into two individual terms $|\Delta^t_i|$ and $|\Delta^t_j|$. Those terms are controlled by the learning rate $\eta^t$ which is convergent as $t$ increases. Then both $|\Delta^t_i|$ and $|\Delta^t_j|$ converges, which means the original series in Eq.~\eqref{eq:series derivation} is convergent. Then the combines series in Eq.~\eqref{eq:error difference} is convergent, which means $\epsilon^T-\epsilon^1$ is convergent while $T\rightarrow\infty$. It means the error $\epsilon^t$ will converge to zero with approximated joint probability.
	\par
	Proof on $\epsilon$-Convergence is a weak condition that can guide us to design our algorithm. In our implementation, a small multiplier is designed to diminish the error produced during approximation. Larger multiplier will push the solution too hard and increase $\epsilon$, which will cause divergence on the derivative that comes from an approximated joint probability. We will discuss the detailed design in the next section.
	
	\subsection{Minimizing Mutual Information for Unsupervised Hash}
	\label{subsec:shuffle & learn}
	Minimizing mutual information can diverge code distribution in binary space. In this section, we will introduce the proposed relaxation method \textit{Shuffle and Learn}. The minimization is served as a shuffling process which encourages the encoder to use up the full binary space. A proper amount of push will guide the model to escape local optimum. We will mainly follow both the network and loss function setup in GreedyHash \cite{su2018greedy} as the shuffling process is independent to the regular unsupervised hashing optimization.
	\par
	To make the update precisely, the full dataset will be used to provide an accurate estimation on probabilities. For a finite dataset, the probability distribution can be treated as an approximation of the realistic distribution. Hence, we need to assume that we have enough samples to the observed probability distribution function $\hat{P}(B_i)$ to be closed to the realistic probability distribution function $P(B_i)$.
	\par
	The proposed approach \textit{Shuffle and Learn} is described as pseudo-code in Algorithm \ref{alg: min mi algorithm}. Shuffling process always happens at the beginning of every epoch, encouraging the network to fill up the whole binary space. It will diverge the code distribution and help to jump out of the local optimum on the learned hash. 
	\par
	A regular optimization on the unsupervised learning process is also needed to collaborate optimization on binary code. In our implementation, we adopted a cosine similarity loss \cite{su2018greedy}  on binary code from the actual learning part in the proposed \textit{Shuffle and Learn} algorithm. It will stimulate the learned binary code to imitate angular relationship on the input feature as shown in Eq.~\eqref{eq:sim loss}. 
	\begin{equation}
	\label{eq:sim loss}
	L_{sim} = \|sim(H_1, H_2) - sim(B_1, B_2)\|_2^2
	\end{equation}
	where $H_1$ and $H_2$ are input feature. $B_1$ and $B_2$ are the hash and the function $sim(\cdot)$ is the cosine similarity function defined in Eq. \eqref{eq: cos sim}.
	\begin{equation}
	\label{eq: cos sim}
	sim(A, B) = \frac{AB}{\|A\|\|B\|}
	\end{equation}
	where $A$ and $B$ are vectors with same number of dimension.
	\par
	Recalling the objective function that optimizes the model, we combined the consistency regularization $L_{reg}$ defined in Eq. \eqref{eq: consistency loss} with the cosine similarity loss $L_{sim}$ with hyper-parameter $\alpha$ as the regular unsupervised hashing loss $L_r$. The consistency loss will align representation between two sides at the binarization and the cosine similarity loss $L_{sim}$ is a regular unsupervised hash learning step. Mutual information loss $L_m$ is updated with all samples collected in training set before every epoch to capture accurate estimation on mutual information. The algorithm is described as Algorithm \ref{alg: min mi algorithm}.
	\begin{equation}
	\label{eq: consistency loss}
	L_{reg} = \|H-B\|_2^2
	\end{equation}
	\begin{algorithm}[ht]
		\SetAlgoLined
		\KwResult{Obtain function parameter $\theta$}
		Initialize network parameter $\theta$, learning rate $\eta$, \\
		\hskip0.2cm hyper-parameter $\alpha$ and $\beta$\;
		\ForEach{epoch}{
			Estimate joint probability $\hat{P}(B_i, B_j)$ for $i,j\in{1..N}$\;
			Calculate mutual information loss $\beta L_{m}$\;
			Update $\theta$ with the approximated derivative $\eta\frac{\partial L_m}{\partial \theta}$\;
			\ForEach{minimatch in dataset}{
				Calculate regular loss $L_{r}$\;
				Update $\theta$ according to $L_r=L_{sim}+\alpha L_{reg}$ \\
				\hskip0.2cm with learning rate $\eta$\;
			}
		}
		\caption{Shuffle and Learn}
		\label{alg: min mi algorithm}
	\end{algorithm}
	\par
	In our implementation, we extract feature from images with a pre-trained VGG-16 \cite{simonyan2014very} without optimizing its parameters and the generated binary representation is collected over the whole dataset to compute the mutual information. Only a naive fully connected layer is adopted to consume the input feature to the learned hash. To stabilize the training, we minimize mutual information only on $P(B_i, B_j)$ and gradients from marginal probability are cut from the back propagation. Also, the mutual information is accumulated as a triangular matrix to stabilize training and the loss function is applied with a multiplier $\beta$ to balance the magnitude on gradients.
	\par
	Notably, the minimization process is concave so that the whole process will not have global solution to the optimization. The convexity on the proposed minimization is not necessary as we adopt this minimization as a relaxation process. Experiments results also suggest that the proposed relaxation is effective on encouraging less code conflict in the binary space.
	\section{Experiments}
	\label{sec:experiments}
	\begin{table*}[ht]
		\centering
		\begin{tabular}{ l | c c c | c c c | c c c }
			\hline
			& \multicolumn{3}{c|}{CIFAR-10} &\multicolumn{3}{|c|}{NUS-WIDE} & \multicolumn{3}{|c}{MS-COCO}\\
			&16 bits &32 bits &64bits &16 bits & 32 bits &64 bits &16 bits &32 bits &64 bits\\
			\hline
			SpherH \cite{heo2012spherical}&0.254&0.291&0.333&0.495&0.558&0.582&0.516&0.547&0.589\\
			ITQ \cite{gong2012iterative}&0.305&0.325&0.349&0.627&0.645&0.664&0.598&0.624&0.648\\
			DGH \cite{liu2014discrete}&0.335&0.353&0.361&0.572&0.607&0.627&0.613&0.631&0.638\\
			DeepBit \cite{lin2016learning}&0.194&0.249&0.277&0.392&0.403&0.429&0.407&0.419&0.430\\
			SGH \cite{dai2017stochastic}&0.435&0.437&0.433&0.593&0.590&0.607&0.594&0.610&0.618\\
			BinGAN \cite{zieba2018bingan}&0.476&0.512&0.520&0.654&0.709&0.713&0.651&0.673&0.696\\
			HashGAN \cite{ghasedi2018unsupervised}&0.447&0.463&0.481&-&-&-&-&-&-\\
			DVB \cite{shen2019unsupervised}&0.403&0.422&0.446&0.604&0.632&0.665&0.570&0.629&0.623\\
			DistillHash \cite{yang2019distillhash}&0.284&0.285&0.288&0.667&0.675&0.677&-&-&-\\
			GreedyHash \cite{su2018greedy}&0.448&0.473&0.501&0.633&0.691&0.731&0.582&0.668&0.710\\
			\hline
			\textit{Shuffle and Learn} (Ours)&\textbf{0.507}&\textbf{0.562}&\textbf{0.592}&\textbf{0.715}&\textbf{0.752}&\textbf{0.777}&\textbf{0.703}&\textbf{0.756}&\textbf{0.789}\\
			\hline
		\end{tabular}
		\caption{\sc MAP Evaluation with Unsupervised Binary Representation}
		\label{tab: numerical}
	\end{table*}
	\begin{figure*}
		\centering
		\begin{subfigure}[b]{0.3\linewidth}
			\centering
			\includegraphics[width=\linewidth]{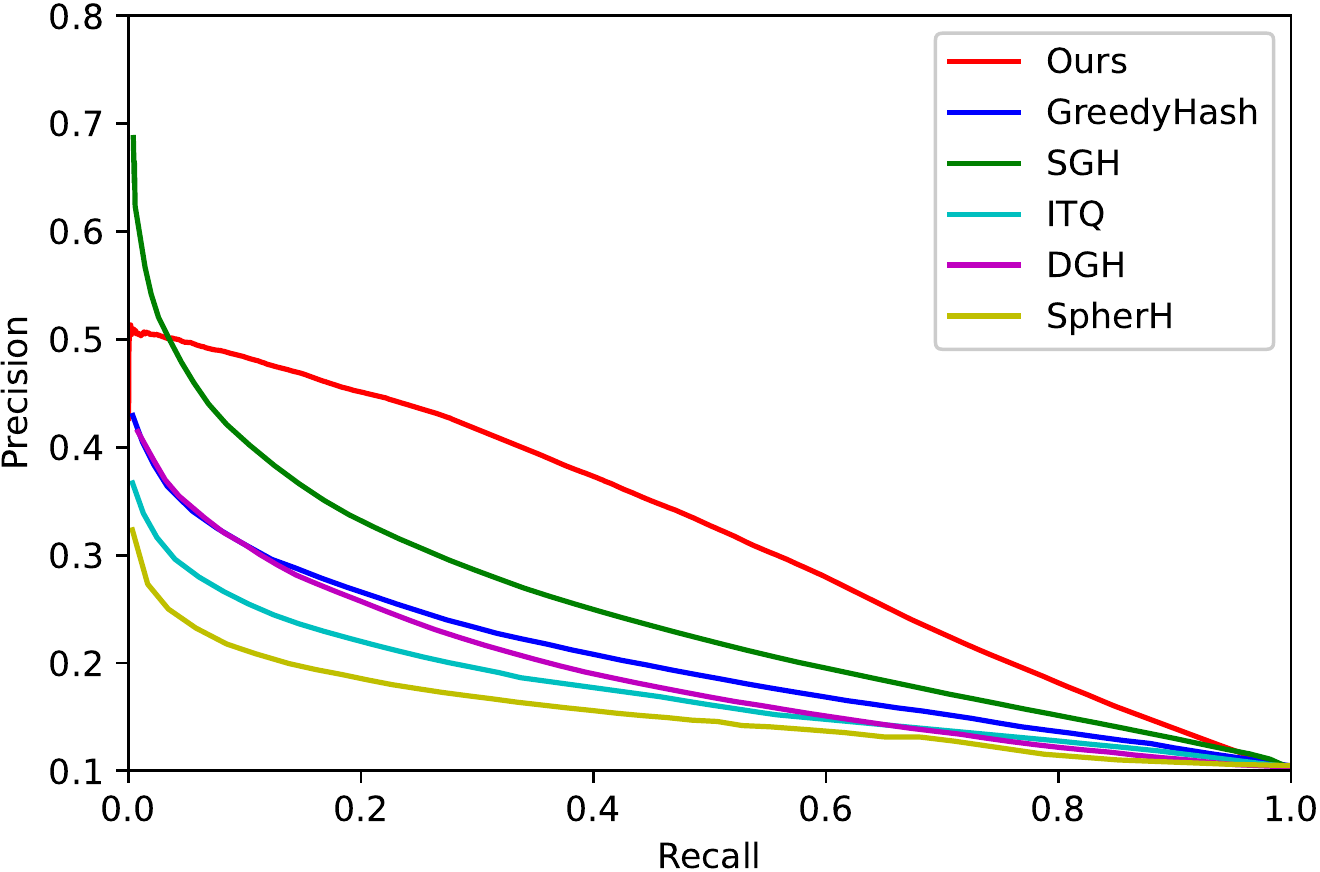}
			\caption{16 Bits}
		\end{subfigure}
		\hskip0.5cm
		\begin{subfigure}[b]{0.3\linewidth}
			\centering
			\includegraphics[width=\linewidth]{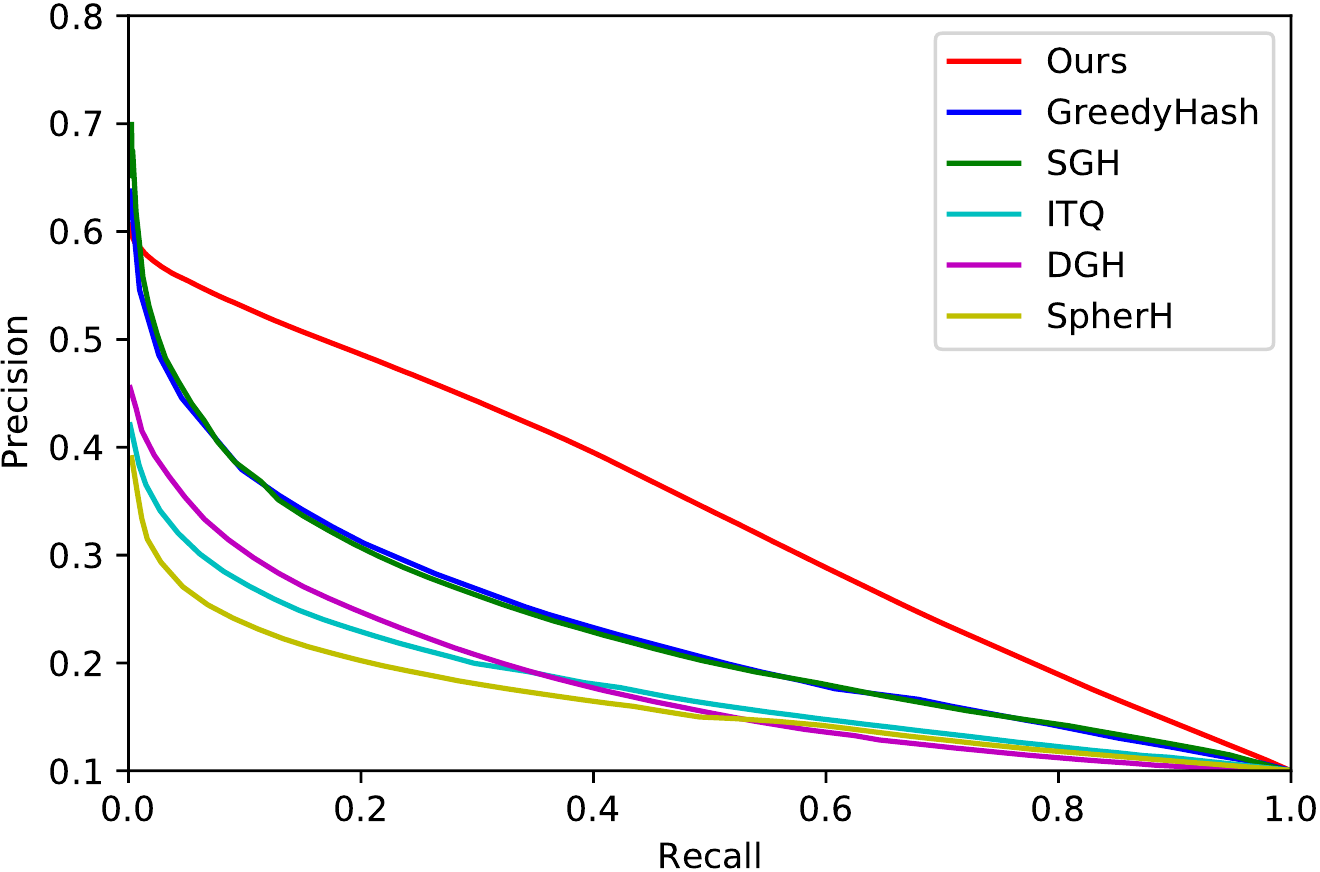}
			\caption{32 Bits}
		\end{subfigure}
		\hskip0.5cm
		\begin{subfigure}[b]{0.3\linewidth}
			\centering
			\includegraphics[width=\linewidth]{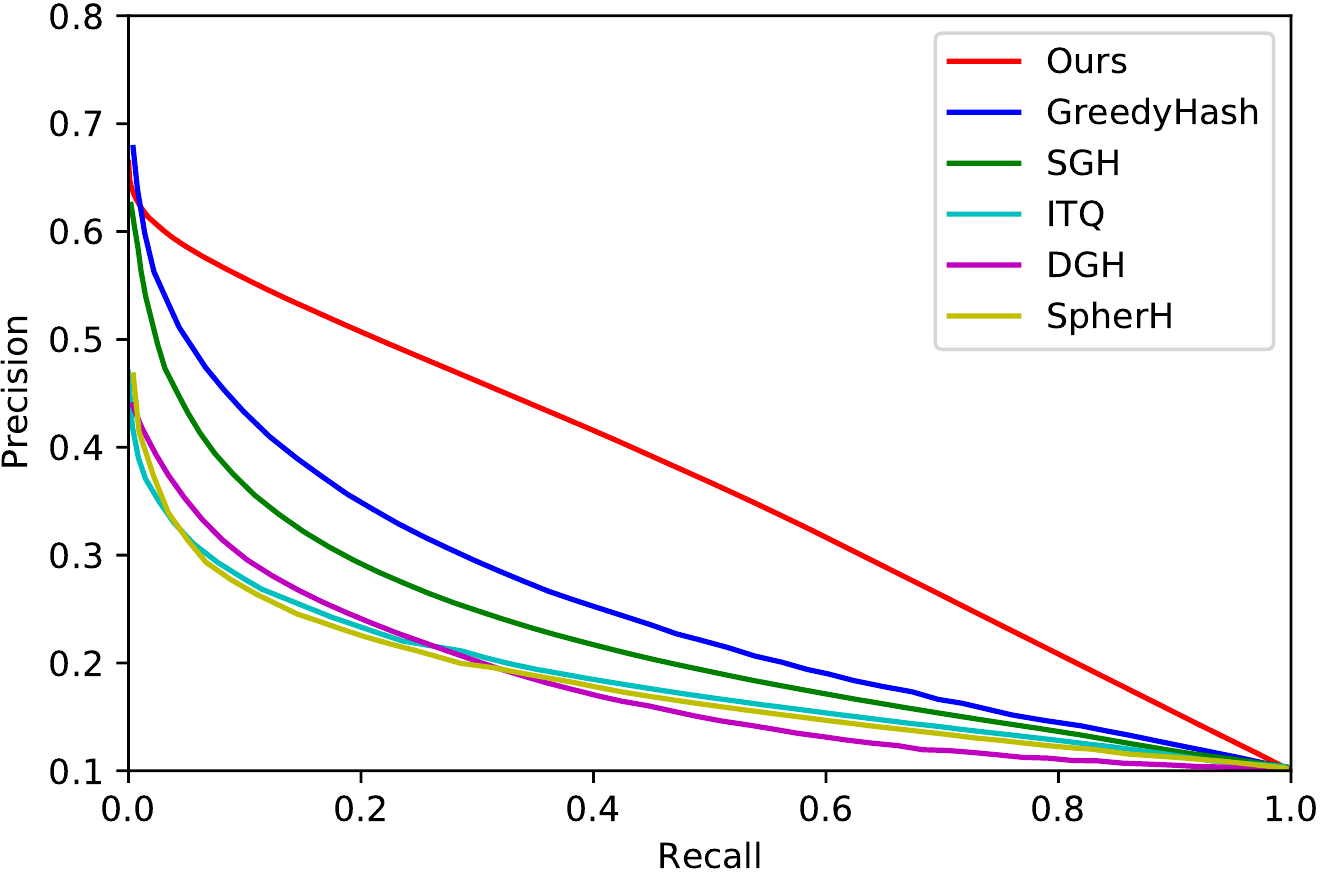}
			\caption{64 Bits}
		\end{subfigure}
		\caption{\sc Precision-Recall Curves of Ours and Compared Methods on CIFAR-10 Dataset}
		\label{fig:pr curve compare cifar}
	\end{figure*}
	
	We conducted experiments to evaluate the performance of the proposed unsupervised hashing algorithm on open datasets. For every experiment in this section, fixed random seed with deterministic behaviour is applied to train the model. We used 0 as our random seed for both network and data loader to improve reproducibility.
	
	\subsection{Datasets}
	The experiments is conducted on three open datasets: CIFAR-10 \cite{krizhevsky2009learning}, NUS-Wide \cite{chua2009nus} and MS-COCO \cite{lin2014microsoft}. We did not apply any data augmentation technique during training and all class labels are not used in our unsupervised setup. All input images are resized to $256\times 256$ before being centre cropped to size $224\times224$ before being fed into the network and are normalized according to mean pixel and standard deviation of pixels on ImageNet \cite{deng2009imagenet}. 
	\begin{enumerate}
		\item \textbf{CIFAR-10}
		{
			is an RGB image dataset with 60K $32\times32$ images with class annotations from 10 different categories. We followed the CIFAR-10 (II) setting in GreedyHash \cite{su2018greedy} which takes 5,000 images each class for training and the rest of 10,000 images for querying. The training dataset will also be served as the retrieval set in this setting. The top-1000 similar images will be considered in our mean average precision (MAP@1000) evaluation.
		}
		\item \textbf{NUS-Wide}
		{
			contains about 270,000 images with 81 different concepts. There can be multiple concepts for a single image. A subset of 21 most common concepts is used in our experiments, picking 195,834 images for the experiment. We followed the data split setup in \cite{xia2014supervised}, taking 500 images from each concept for training and 100 images each category for querying. The rest of the data are kept as retrieval dataset, providing similar samples for every query image in the test set. We adopted the same setting as other work \cite{xia2014supervised}, taking top-5000 neighbours to evaluate MAP on NUS-Wide.
		}
		\item \textbf{MS-COCO}
		{
			provides images from a large scope of concepts. In our experiments we used \textit{trainval2014} for COCO dataset to match the setup in HashNet \cite{cao2017hashnet}. The pruned dataset contains 12,2218 images from 80 different classes. 5,000 images are randomly selected for query and another 10,000 are also picked for training purpose. The remaining is reserved as database during MAP evaluation. We follow common evaluation setting \cite{cao2017hashnet,zhu2016deep} on COCO dataset, considering top-5000 similar images to accumulate MAP score.
		}
	\end{enumerate}

	\begin{figure*}[ht]
	\centering
	\includegraphics[width=0.85\linewidth]{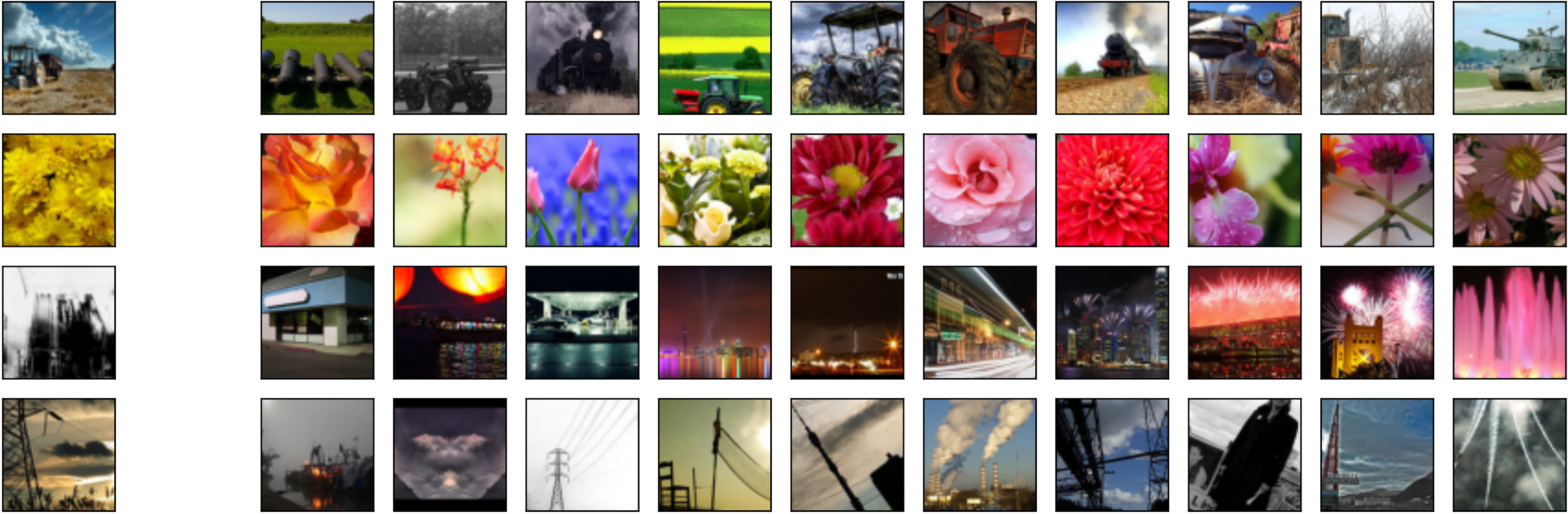}
	\caption{\sc Retrieved Images on Right with Query Image on Left from NUS-Wide with 16 Bits }
	\label{fig:retrieved 16}
	\end{figure*}
	
	\subsection{Experimental Setup}
	Experiments are conducted using a batch size of 32 and a learning rate of $1e-3$. We set hyper-parameters $\alpha=0.1$ in our experiments and $1e-4$, $1e-3$, $1e-2$ on $\beta$ for our 16-bits, 32-bits, 64 bits model respectively. A multi-step learning rate decay is applied in our experiments. It will decay by a rate of 0.1 every 100 epoch. We trained the network for 300 epochs with standard SGD optimizer with the momentum of 0.9 and weight decay of $5e-4$. Notably, a special optimizer is applied to mutual information minimization, which does not have any momentum or weight decay. This strategy can ensure that the network will only take the precise gradient to shuffle the binary representation.
	
	\subsection{Evaluation Metrics}
	Mean average precision (MAP) is used to evaluate the model's performance on Image retrieval. It is widely used in retrieval evaluations \cite{heo2012spherical,gong2012iterative,liu2014discrete,lin2016learning,dai2017stochastic,zieba2018bingan,ghasedi2018unsupervised,shen2019unsupervised,yang2019distillhash,su2018greedy}. Average precision(AP) is obtained by accumulating retrieval precision on each classes. AP will decrease as we increase the number of retrieved images. Mean of average precision will average all collected average precision score, which is a good criterion for evaluation over different retrieval setups.
	\par
	Precision recall curve will present the performance with more visual details. It illustrates the relation between precision and recall, which is another form to describe retrieval performance over different settings. Notably, the integral of the precision-recall curve is positively related to MAP score.

	\subsection{Evaluation Results}
	\subsubsection{Benchmark on Image Retrieval}
	Evaluation results on MAP are illustrated in Table \ref{tab: numerical}. All methods compared are using the same deep VGG-16 feature to ensure a fair comparison. Identical training and test split setup are also used to evaluate the proposed method.
	\par
	We compared the proposed algorithm with several state-of-the-art methods including SpherH \cite{heo2012spherical}, ITQ \cite{gong2012iterative}, DGH \cite{liu2014discrete}, DeepBit \cite{lin2016learning}, SGH \cite{dai2017stochastic}, BinGAN \cite{zieba2018bingan}, HashGAN \cite{ghasedi2018unsupervised}, DVB \cite{shen2019unsupervised}, DistillHash \cite{yang2019distillhash}, GreedyHash \cite{su2018greedy}. According to evaluation results on image retrieval in Table \ref{tab: numerical}, the proposed method has advantages on all code length settings comparing to current state-of-the-art methods. Large gaps can be observed on 64-bits with the proposed \textit{Shuffle and Learn} algorithm. The proposed regular optimization step is similar to GreedyHash \cite{su2018greedy} as they both use cosine similarity loss and the same regularization loss during training. It can be observed that the mutual information minimization did help the network to improve representativity over the whole domain. Even comparing to the generative method like BinGAN \cite{zieba2018bingan} and HashGAN \cite{ghasedi2018unsupervised}, retrieval accuracy of the proposed method can still over-perform them with considerable improvement.
	\par
	The proposed \textit{Shuffle and Learn} achieves higher performance on larger code length. Results with 64 bits code on CIFAR-10(II) achieved 59.2\% on MAP score which is 9.1\% higher than GreedyHash \cite{su2018greedy} and 7.2\% higher than BinGAN \cite{zieba2018bingan}. The improvement made on NUS-Wide and MS-COCO is 4.5\% and 7.5\% comparing to the highest score among SOTA. It suggests that code conflict is indeed a challenge to high-quality hashing.  And also proper shuffling on code distribution can significantly improve hashing quality in unsupervised setups.
	\par
	Precision-Recall curves are also collected to compare the proposed approach to others. We used CIFAR-10 and $\beta$ is set as described in Section \ref{subsec:shuffle & learn}. Results are demonstrated in Fig. \ref{fig:pr curve compare cifar}. Most compared methods cannot retrieve semantically related clusters from the database, which means those hashing approaches may not be able to generalize semantic hash for similar samples. Those methods surely provide accurate neighbours but are trapped in local optimum with no constraint on code generation. However, with the proposed mutual information, precision is well kept at low recall, which means the hashing model trained with mutual information loss can be more robust and informative on clustering semantically related samples. 
	
	\subsubsection{Visualization}
\begin{figure}
		\centering
		\begin{subfigure}[b]{0.47\linewidth}
			\centering
			\includegraphics[width=\linewidth]{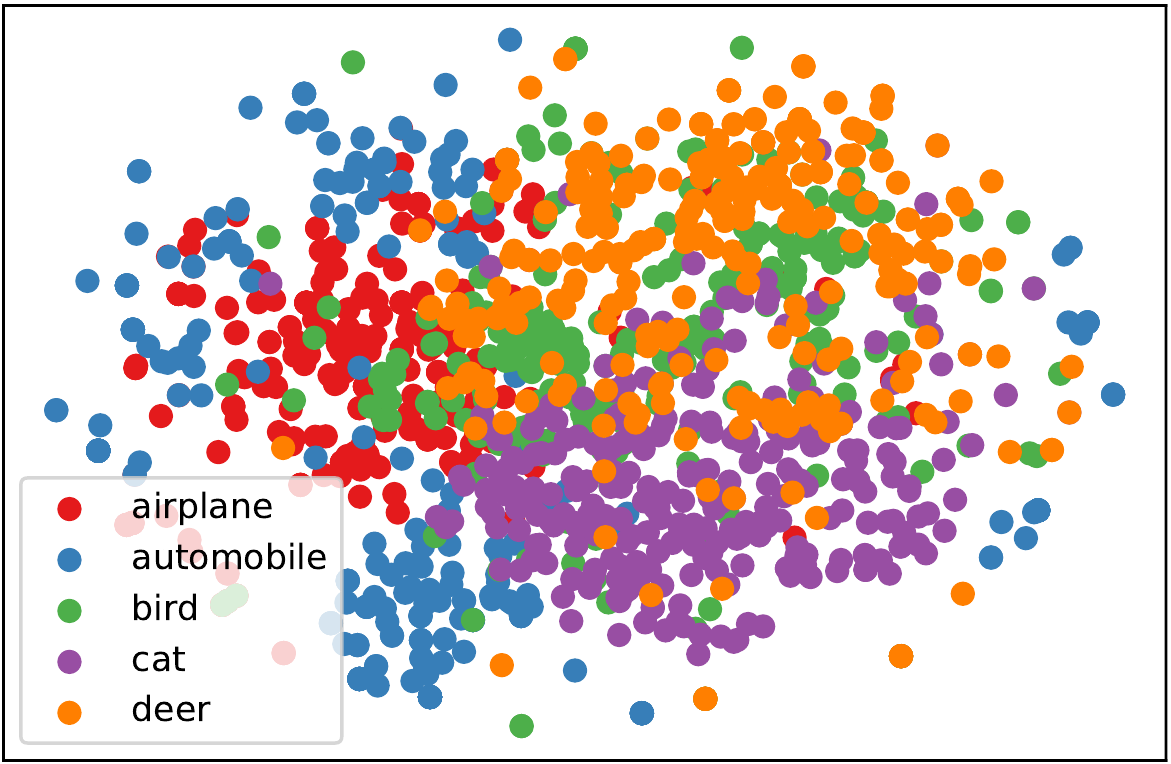}
			\caption{32 Bits}
		\end{subfigure}
		\hskip0.2cm
		\begin{subfigure}[b]{0.47\linewidth}
			\centering
			\includegraphics[width=\linewidth]{{tSNE_64}.pdf}
			\caption{64 Bits}
		\end{subfigure}
		\caption{\sc t-SNE Visualization on CIFAR-10 Dataset}
		\label{fig:t-sne}
	\end{figure}

	We adopted t-SNE \cite{maaten2008visualizing} to visualize code quality on the learned binary representation. Results on visualization with our 32 bits and 64 bits hash is illustrated in Fig. \ref{fig:t-sne}. The unsupervised binarized representation can separate samples according their semantic contents. And even with smaller form, for example, binary representation with only 32 bits, can till gather samples as neighbours concerning the semantic label without any supervision from the ground truth. 
	\par
	We also visualized the retrieved images with the query image in Fig. \ref{fig:retrieved 16}. We randomly select 4 query images and retrieved 10 most similar images from the database set of NUS-Wide. The top-10 nearest neighbours are semantically related, which suggest that the proposed relaxation can help a simple model to retrieve semantic neighbours more effectively with the unsupervised binary hash.
	
	\begin{figure*}[h]
		\centering
		\begin{subfigure}[b]{0.3\linewidth}
			\centering
			\includegraphics[width=\linewidth]{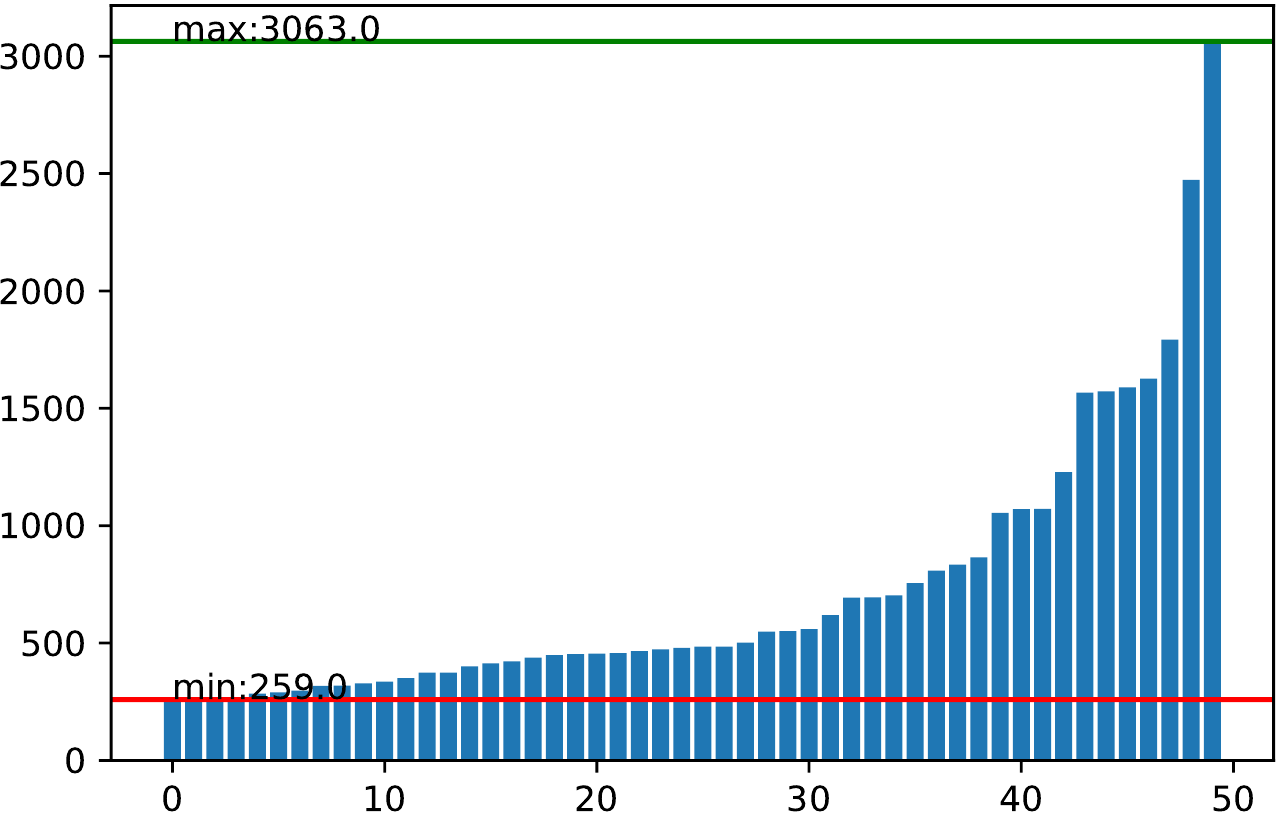}
			\caption{$\beta=0.0001$}
		\end{subfigure}
		\begin{subfigure}[b]{0.3\linewidth}
			\centering
			\includegraphics[width=\linewidth]{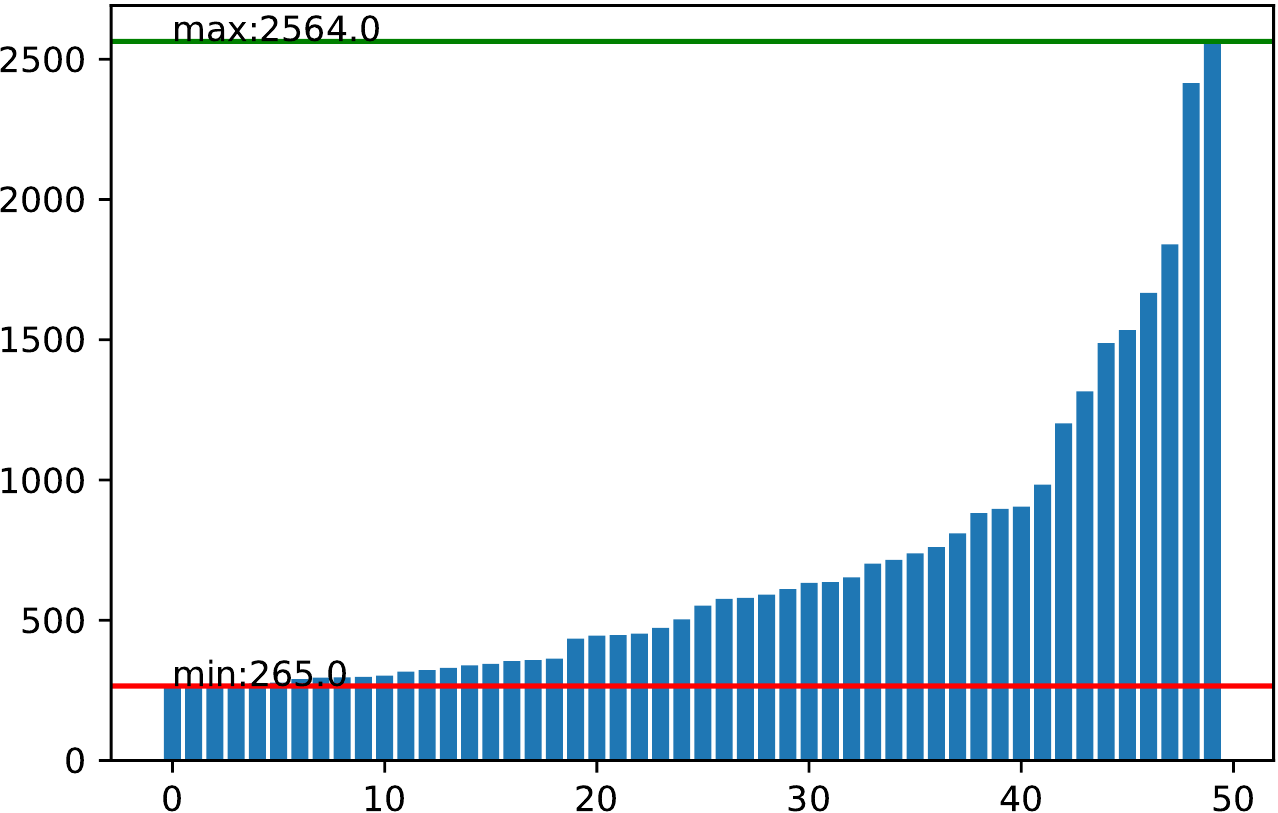}
			\caption{$\beta=0.001$}
		\end{subfigure}
		\begin{subfigure}[b]{0.3\linewidth}
			\centering
			\includegraphics[width=\linewidth]{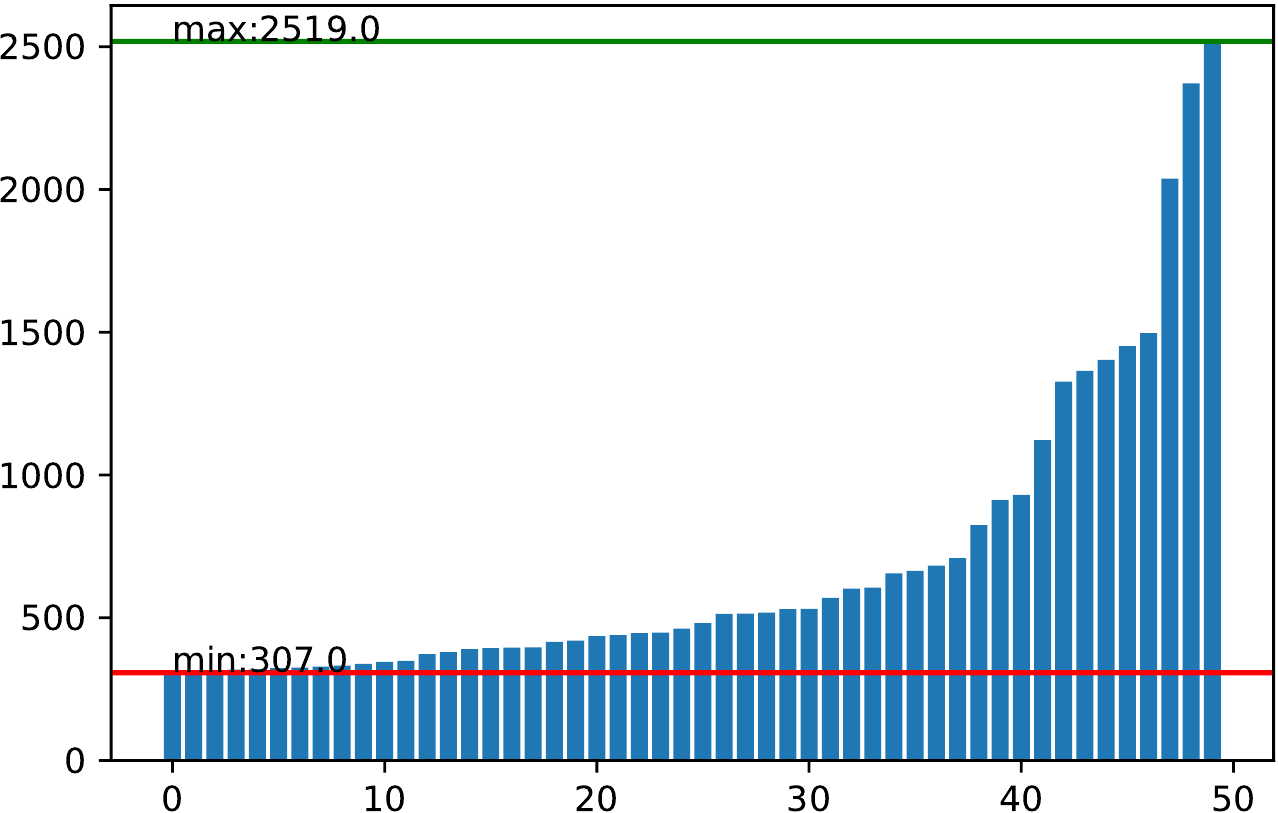}
			\caption{$\beta=0.01$}
		\end{subfigure}
		\caption{\sc Binary Space Utilization under Different $\beta$}
		\label{fig:code space comp}
	\end{figure*}
	
	\subsection{Empirical Analysis}
	\subsubsection{Ablation Study}
	According to our assumption, a proper amount of pushing can improve performance on hashing. Hyper-parameter $\beta$ is designed to control the strength of push during optimization. A smaller value for $\beta$ will diminish the effect of minimizing mutual information while larger $\beta$ will degrade the code quality in performance on retrieval.
	\begin{table}[h]
		\caption{\sc MAP Results on Models Trained with Different $\beta$}
		\centering
		\begin{tabular}{ c | c c c c c c}
			\hline
			$\beta$&0.1&0.01&0.001&0.0001&0.00001&0\\
			\hline
			Ours 16 bits &0.465&0.474&0.482&\textbf{0.507}&0.493&0.477\\
			Ours 32 bits &0.539&0.553&\textbf{0.562}&0.554&0.560&0.549\\
			Ours 64 bits &0.591&\textbf{0.592}&0.591&0.590&0.589&0.589\\
			\hline
		\end{tabular}
		\label{tab:beta ablation}
	\end{table}
	\begin{figure}[h]
		\centering
		\includegraphics[width=0.6\linewidth]{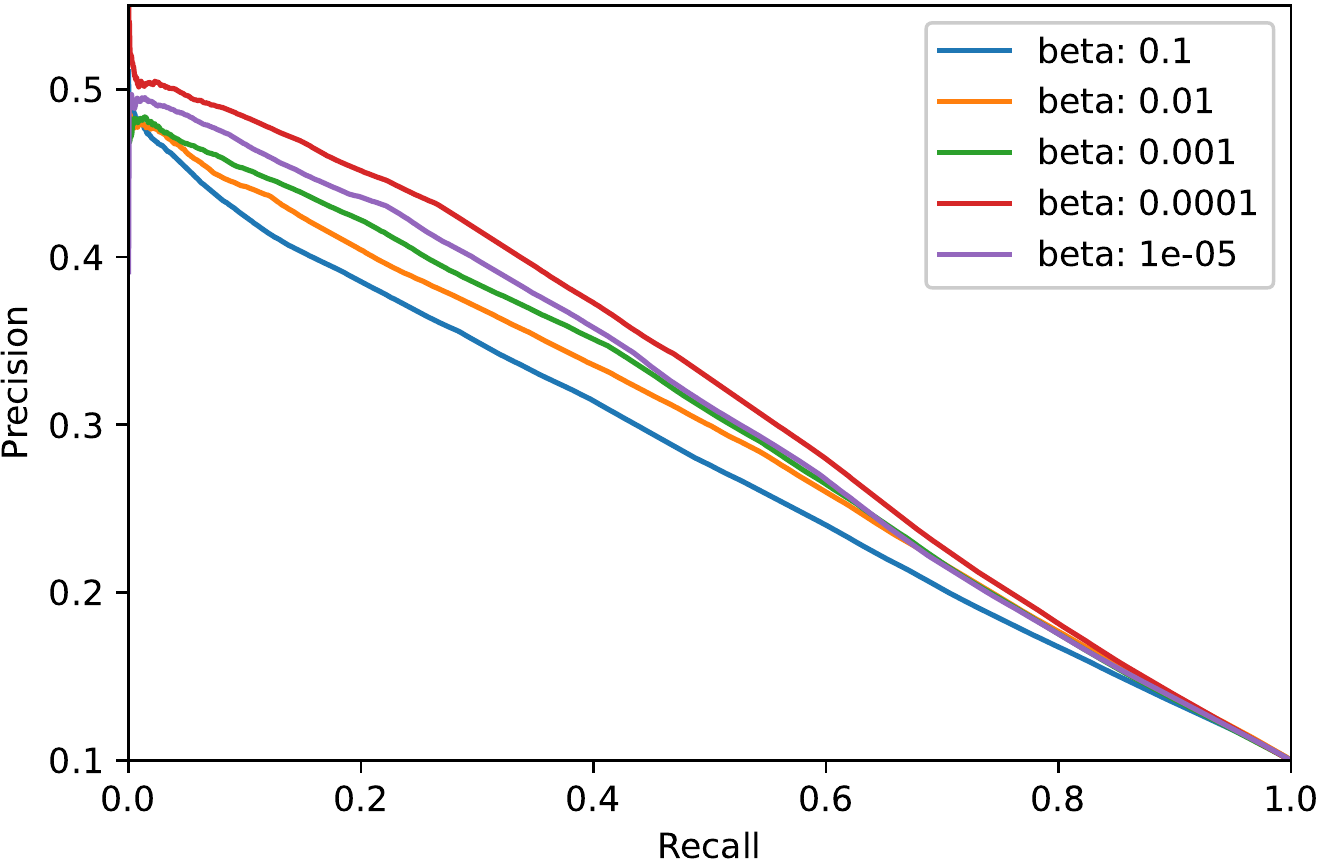}
		\caption{\sc Precision-Recall Curve with Different $\beta$ on 16 Bits}
		\label{fig:pr curve beta}
	\end{figure}
	\par
	In Table \ref{tab:beta ablation}, we demonstrated our controlled experiment with hyper-parameter $\beta$ on CIFAR-10(II) dataset. Results are evaluated on MAP and every setup is the same except $\beta$. Table \ref{tab:beta ablation} proved our assumption on $\beta$, showing us that appropriate push is needed to improve the code quality. A larger $\beta$, for example $\beta=0.1$ will disrupt the regular unsupervised training as it leads to a lower the performance that is even worse than the baseline($\beta=0$). Also, if $\beta$ is set to a too small value, eg. $\beta=1e-5$, the optimization will degrade to baseline method. Precision-Recall curve with different $\beta$ values on 16 bits setup in Fig. \ref{fig:pr curve beta} also gives a strong evidence on the previous assumption. Overall code quality is incrementally improved as we move closer to the optimal value on $\beta$.
	\par
	On the other hand, the optimal value for $\beta$ increases as the code length goes up. Smaller code length will cause more accumulative gradients on each bit so that the mutual information will be more dominant compared to the regular unsupervised constraints. A larger code length could help the gradients to relax by distributing the error to other bits. But still, the proper $\beta$ is crucial to encourage the binary code to fill up the full space.
	
	\subsubsection{Code Analysis}
	We collected statistics to evaluate how will minimization on mutual information would affect the code utilization in binary space. We controlled hyper-parameter $\beta$ to assess the binary space utilization with different setups. We used CIFAR-10 dataset and follows the same training protocol as described above. The code statistics are sorted according to the counted number to evaluate the utilization of binary space.
	\par
	As shown in Fig \ref{fig:code space comp}, minimization on mutual information can encourage the network to use more keys in the binary space. The maximal value on single key decreases as we increase the value of $\beta$. Also, the minimum of the code count increases which means the code distribution is more flattened. It can be interpreted as more binary space is used as we shuffle harder by minimizing the mutual information with the proposed method. Though it is achieving what we desired, it does not mean larger $\beta$ is good for a hashing algorithm. According to the conducted ablation study in the previous section, larger $\beta$ will cause performance drop as it may shuffle the code too hard when searching for generalized semantic hash in the binary space.

	\subsubsection{Effect of Mutual Information Minimization on Binary Code}
\begin{figure}[h]
		\centering
		\begin{subfigure}[b]{0.45\linewidth}
			\centering
			\includegraphics[width=\linewidth]{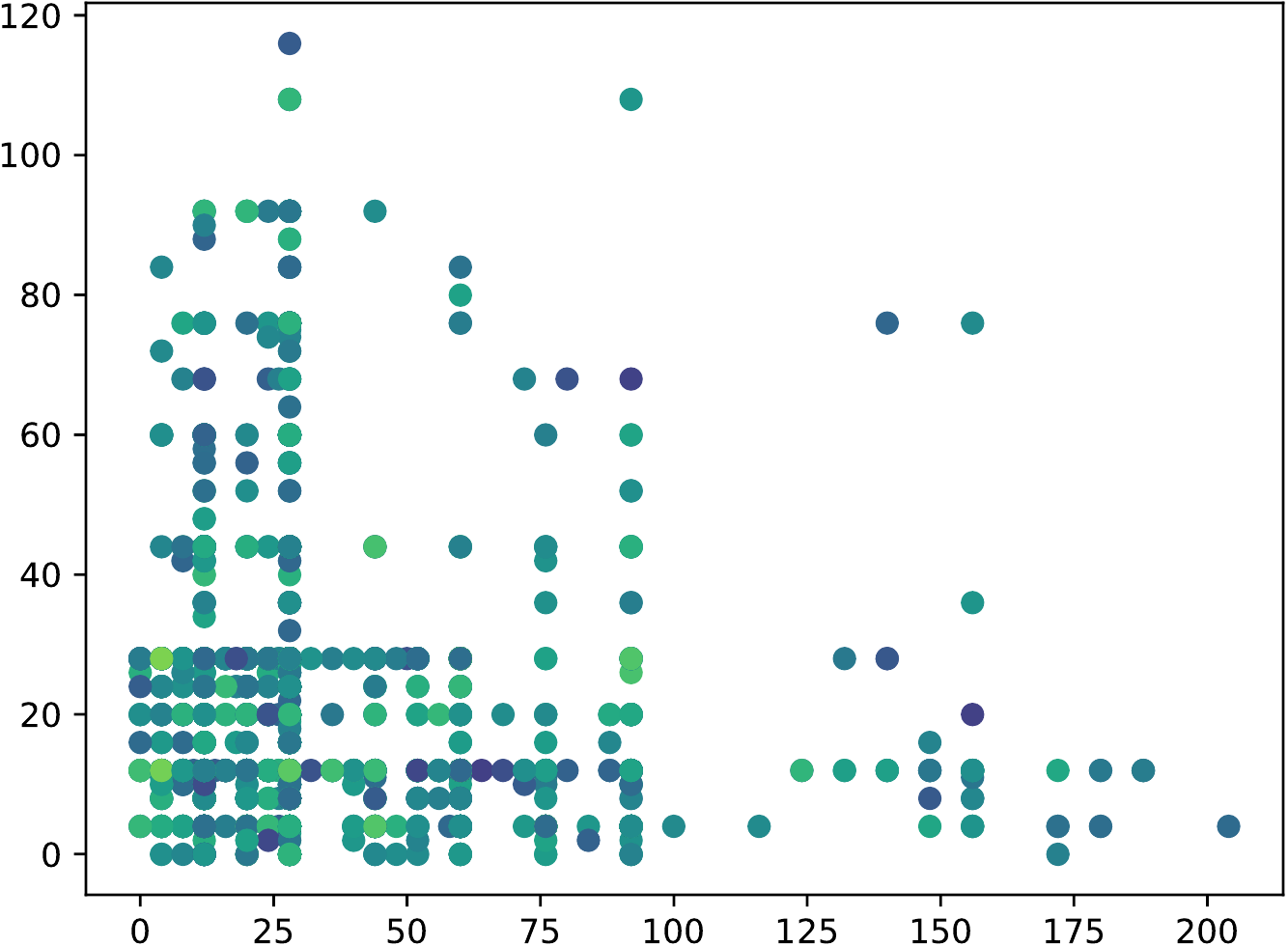}
			\caption{Step 0}
		\end{subfigure}
		\begin{subfigure}[b]{0.45\linewidth}
			\centering
			\includegraphics[width=\linewidth]{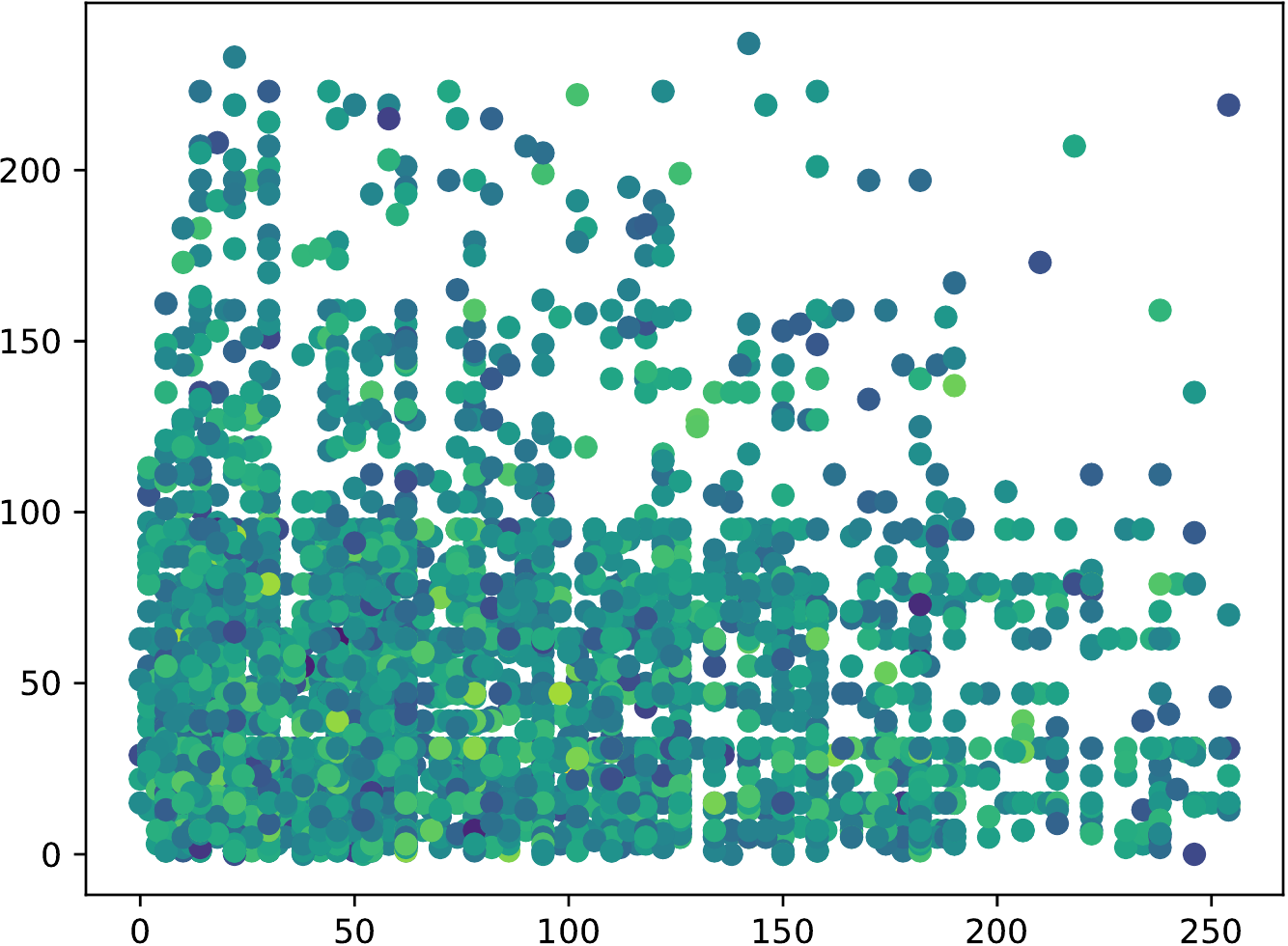}
			\caption{Step 1}
		\end{subfigure}
		\begin{subfigure}[b]{0.45\linewidth}
			\centering
			\includegraphics[width=\linewidth]{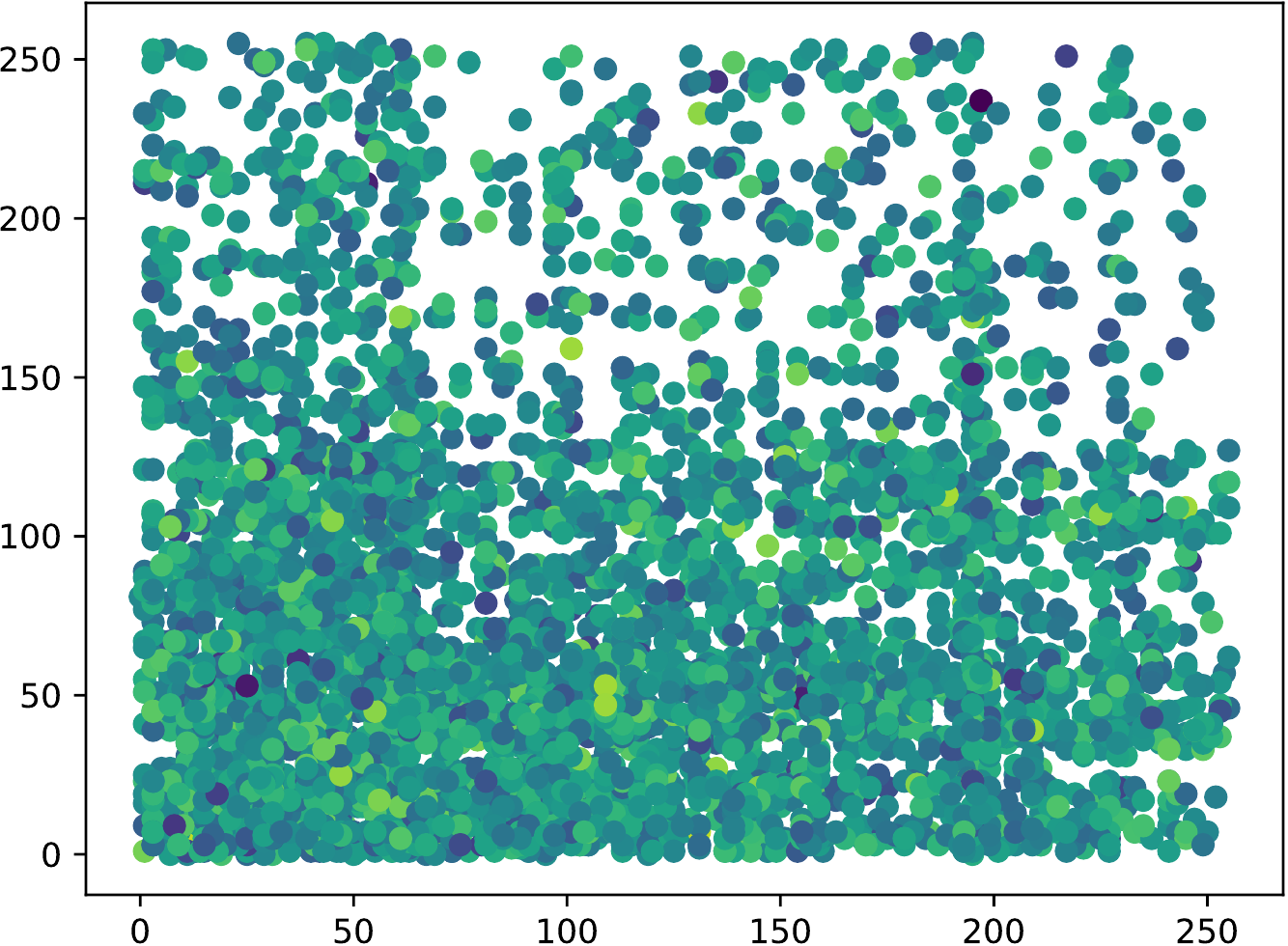}
			\caption{Step 2}
		\end{subfigure}
		\begin{subfigure}[b]{0.45\linewidth}
			\centering
			\includegraphics[width=\linewidth]{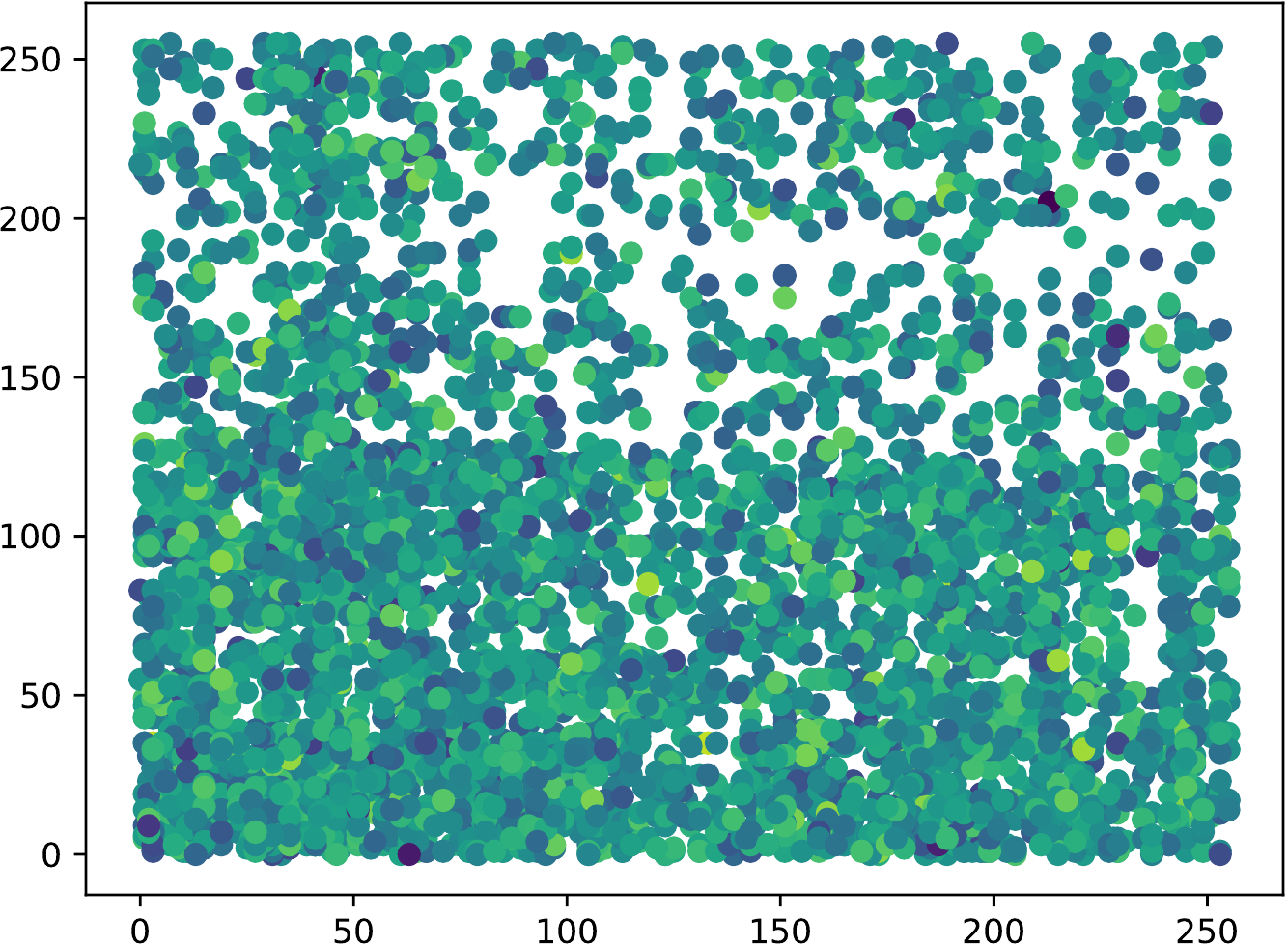}
			\caption{Step 3}
		\end{subfigure}
		\caption{\sc Minimizing Mutual Information over Binary Representation}
		\label{fig:shuffle}
	\end{figure}

	We investigated effects on the proposed algorithm with approximated joint probability during optimization to support the proof on its $\epsilon$-Convergence discussed in Section \ref{subsec:e-convergence}. First, we conducted this experiment with CIFAR-10 (II) dataset, trying to encourage the binary code to utilize the whole binary space. The network will be only optimized according to the mutual information loss on binary code. To visualize, the embedding is divided into two subsets and converted into integers respectively. Then we used two converted integers to represent the sample's coordinates on a 2D space. The visualization result will not demonstrate the semantic relationship but optimization process which justifies our motivation. Result is demonstrated in Fig. \ref{fig:shuffle}.
	\par
	
	We train the network with a learning rate of $1e-5$ with only the optimization step on mutual information loss. The algorithm converges in about 30 iterations. We visualized the binary representation using the strategy discussed above. The binary code is diverging into the full space by minimizing mutual information with the proposed approach. The solution is stable after it converged, which verifies our motivation and design discussed in Section \ref{sec:method}. There will be an optimal solution for minimizing mutual information with the joint probability and the optimization with approximated derivatives of joint probability will converge as $\epsilon$ converges. With this condition, we can ensure that proper minimization on mutual information will not disturb the regular optimization on unsupervised binary representation learning.

	% TODO: Error Analysis on Approximated Joint Probability
	
	\section{Conclusion}
	\label{sec:conclusion}
	
	% TODO: Conclusion:
	
	% Contribution
	% - Global contrib
	%   - Identify the problem
	%   - Then we verify the assumption
	% - Technical 
	% Advantage of this paper
	% Method name - target
	% Why to do the minimization? 
	% mention the experiments? what has been evaluated?
	% 
	% method name repeat
	
	Binary representation has fewer coding space than the continuous and there is a higher chance when two identically different samples share a same code for binary representations. This is often recognized as hash conflict or code conflict in binary space. In this paper, we identified code conflict in low dimensional space as a new barrier to high-quality unsupervised hashing. A novel meta-algorithm called \textit{Shuffle and Learn} is introduced to diverge code distribution in binary space which can enhance hash quality and also help the model escape local minimum. The paper also provides proof on the $\epsilon$-Convergence of the proposed algorithm. Mutual information loss on binary representation can encourage the network to fully utilize the binary space, which mitigates hash conflict on semantically dissimilar samples. With a proper amount of shuffling, the network can jump out of local minimum and also generalize better with the mutual information loss. The proposed approach is flexible and can be applied to other hashing algorithms to enhance model generality. 
	
	Future research should consider studies on a more concrete condition on convergence. More specifically, both theoretical and numerical study is necessary to investigate. Furthermore, current estimation on joint probability is to heavy to perform on very large databases. A flexible on-the-fly technique on the joint probability estimation will broaden its application and also accelerate the training possibility. Minimizing mutual information on continuous outputs is also an interesting direction as it will encourage independence among the output neurons. Learning more identical neurons will eliminate redundancy in networks and encourage more node to be pruned when reducing the size of neural networks.

	% if have a single appendix:
	%\appendix[Proof of the Zonklar Equations]
	% or
	%\appendix % for no appendix heading
	% do not use \section anymore after \appendix, only \section*
	% is possibly needed
	
	% use appendices with more than one appendix
	% then use \section to start each appendix
	% you must declare a \section before using any
	% \subsection or using \label (\appendices by itself
	% starts a section numbered zero.)
	%

	% use section* for acknowledgment

	% Can use something like this to put references on a page
	% by themselves when using endfloat and the captionsoff option.
	\ifCLASSOPTIONcaptionsoff
	\newpage
	\fi

	% trigger a \newpage just before the given reference
	% number - used to balance the columns on the last page
	% adjust value as needed - may need to be readjusted if
	% the document is modified later
	%\IEEEtriggeratref{8}
	% The "triggered" command can be changed if desired:
	%\IEEEtriggercmd{\enlargethispage{-5in}}
	
	% references section
	
	% can use a bibliography generated by BibTeX as a .bbl file
	% BibTeX documentation can be easily obtained at:
	% http://mirror.ctan.org/biblio/bibtex/contrib/doc/
	% The IEEEtran BibTeX style support page is at:
	% http://www.michaelshell.org/tex/ieeetran/bibtex/
	\bibliographystyle{IEEEtran}
	% argument is your BibTeX string definitions and bibliography database(s)
	\bibliography{IEEEabrv,paper}
	%
	% <OR> manually copy in the resultant .bbl file
	% set second argument of \begin to the number of references
	% (used to reserve space for the reference number labels box)

	% biography section
	% 
	% If you have an EPS/PDF photo (graphicx package needed) extra braces are
	% needed around the contents of the optional argument to biography to prevent
	% the LaTeX parser from getting confused when it sees the complicated
	% \includegraphics command within an optional argument. (You could create
	% your own custom macro containing the \includegraphics command to make things
	% simpler here.)
	%\begin{IEEEbiography}[{\includegraphics[width=1in,height=1.25in,clip,keepaspectratio]{mshell}}]{Michael Shell}
	% or if you just want to reserve a space for a photo:
	
	% You can push biographies down or up by placing
	% a \vfill before or after them. The appropriate
	% use of \vfill depends on what kind of text is
	% on the last page and whether or not the columns
	% are being equalized.
	
	%\vfill
	
	% Can be used to pull up biographies so that the bottom of the last one
	% is flush with the other column.
	%\enlargethispage{-5in}

	% that's all folks
\end{document}